\documentclass[letterpaper, 10 pt, conference]{ieeeconf}  

\IEEEoverridecommandlockouts                              

\usepackage{cite}
\usepackage{amsmath,amssymb,amsfonts}
\usepackage{algorithmic}
\usepackage{graphicx}
\usepackage{subfigure}
\usepackage{booktabs}
\usepackage{textcomp}
\usepackage{xcolor}
\usepackage{array}
\usepackage{multirow}
\usepackage{epsfig}
\usepackage{tabularx}
\usepackage{hyperref}
\title{\LARGE \bf ThermoStereoRT: Thermal Stereo Matching in Real Time via Knowledge Distillation and Attention-based Refinement
}

\author {
    Anning Hu\textsuperscript{\rm 1},
    Ang Li\textsuperscript{\rm 1},
    Xirui Jin\textsuperscript{\rm 1},
    Danping Zou\textsuperscript{\rm 1*}%
\thanks{
    $^{1}$Shanghai Key Laboratory of Navigation and Location-based Service,
Shanghai Jiao Tong University. \{huanning, liang\_sjtu, jinxirui\}@sjtu.edu.cn. $*$Corresponding author:(dpzou@sjtu.edu.cn). This work was supported by National Key R\&D Program of China (2022YFB3903801) and National Science Foundation of China (62073214).
}%
}
\begin{document}

\maketitle
\thispagestyle{empty}
\pagestyle{empty}
\begin{abstract}
We introduce ThermoStereoRT, a real-time thermal stereo matching method designed for all-weather conditions that recovers disparity from two rectified thermal stereo images, envisioning applications such as night-time drone surveillance or under-bed cleaning robots. Leveraging a lightweight yet powerful backbone, ThermoStereoRT constructs a 3D cost volume from thermal images and employs multi-scale attention mechanisms to produce an initial disparity map. To refine this map, we design a novel channel and spatial attention module. Addressing the challenge of sparse ground truth data in thermal imagery, we utilize knowledge distillation to boost performance without increasing computational demands. Comprehensive evaluations on multiple datasets demonstrate that ThermoStereoRT delivers both real-time capacity and robust accuracy, making it a promising solution for real-world deployment in various challenging environments. Our code will be released on \url{https://github.com/SJTU-ViSYS-team/ThermoStereoRT}.


\end{abstract}


\section{Introduction}

Stereo matching is a fundamental visual task in robotics\cite{nalpantidis2010stereo}, autonomous driving, and 3D reconstruction\cite{geiger2011stereoscan}. The goal of stereo matching tasks is to determine the disparity between a pair of images captured by two rectified cameras, enabling the reconstruction of depth information. Most stereo matching works concentrate on RGB image pairs\cite{chang2018pyramid,tankovich2021hitnet,lipson2021raft,li2022practical}, however, RGB cameras are prone to being affected by lighting conditions and struggle to operate efficiently in smoky or low-light\cite{sharma2020nighttime} environments.

Thermal imaging cameras\cite{gade2014thermal_survey}, on the other hand, are barely influenced by ambient illumination and thus can function effectively in conditions where RGB cameras fall short, such as foggy\cite{velazquez2022thermal_analysis} or poorly illuminated scenes. With the cost of thermal cameras decreasing, these devices are finding more opportunities for application. However, thermal images often lack texture, are noisier, and tend to have lower resolutions, which poses significant challenges for stereo matching. Additionally, the scarcity of real-world stereo thermal datasets and the absence of synthetic ones make developing robust and accurate thermal stereo matching algorithms challenging.

\begin{figure}[t]
\centering
{\includegraphics[width=8cm]{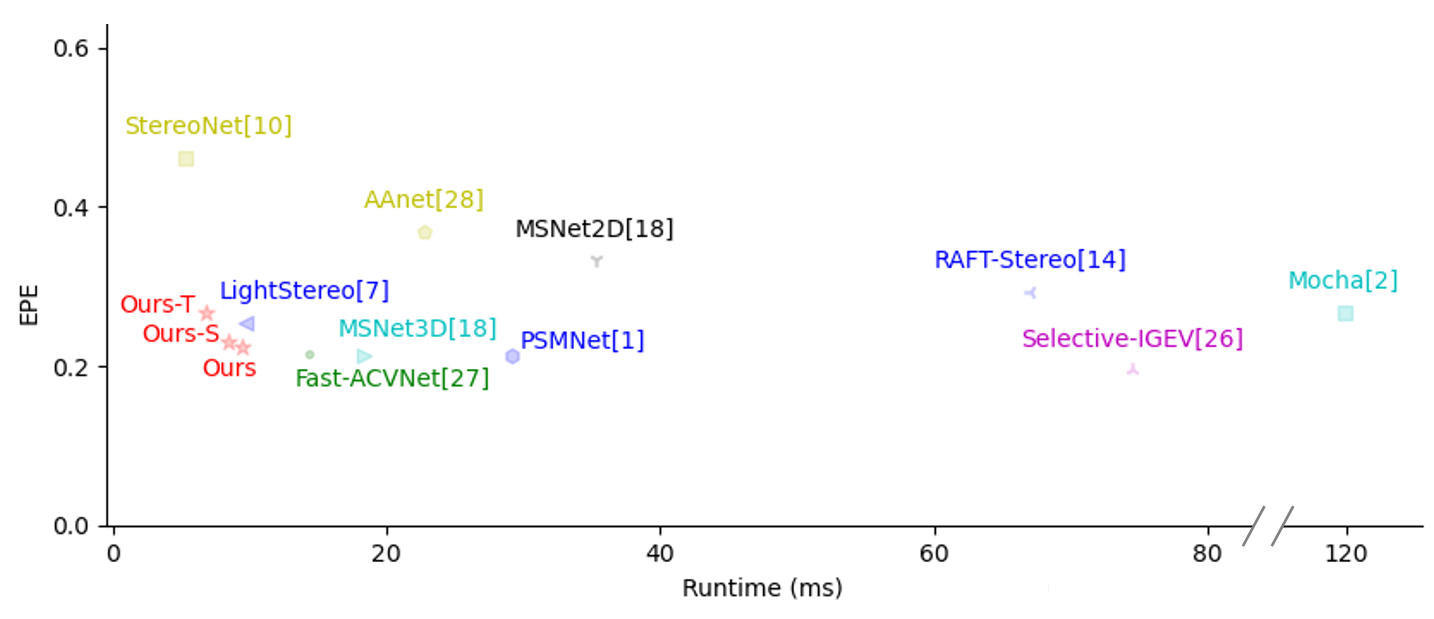}} 
\caption{Our method achieves the best trade off in accuracy and inference
speed on the MS2\cite{shin2023MS2} dataset.}
\label{fig:runtime}
\end{figure}

\begin{figure}[h]
\centering
{\includegraphics[width=0.5\textwidth]{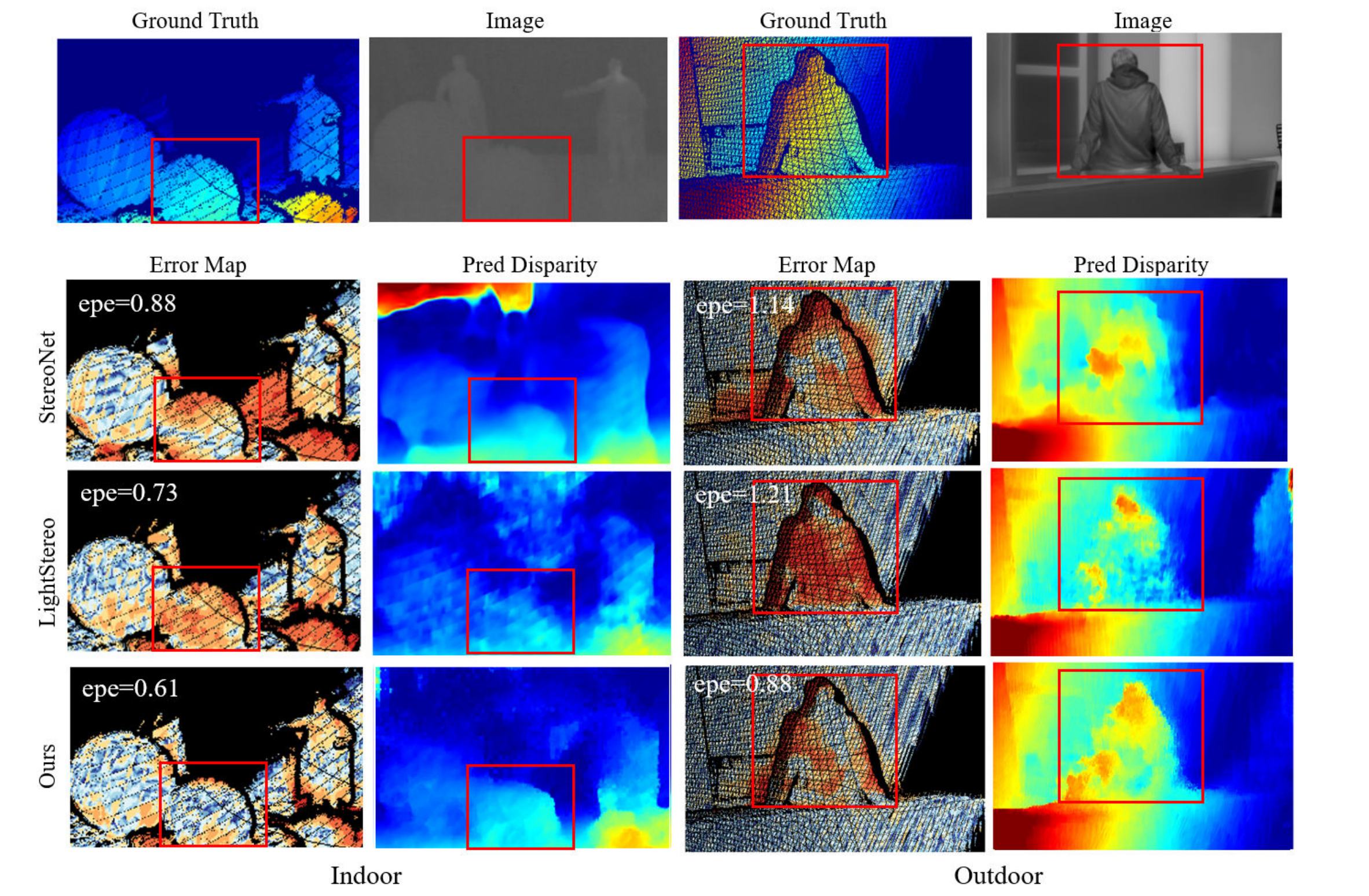}} 
\vspace{-0.5cm}
\caption{Results in both indoor and outdoor scenarios of CATS\cite{treible2017cats} dataset. Our method produces more accurate predictions with smaller disparity errors and more regular object shapes.}
\label{fig:cats}
\vspace{-0.8cm}
\end{figure}

In this work, we propose ThermoStereoRT, a novel real-time thermal stereo matching algorithm that balances accuracy and inference speed. Our method employs a shallow encoder to extract features from left and right thermal images, and the features are used to construct the cost volume. We then apply regression on the cost volume using residual structures and SE\cite{hu2018squeeze} modules, inspired by MobileNetV3\cite{koonce2021mobilenetv3}, to enhance multi-scale channel features with attention, producing an initial disparity map. We further enhance the channel and spatial features of the left image to derive feature attention weights, which are used to refine the initial disparity map at multiple scales, producing the final, detailed disparity map. All components employ lightweight operations to ensure real-time performance.

To address the challenges from limited datasets\cite{shin2023MS2, treible2017cats} and sparse ground truth in thermal stereo matching, we employ knowledge distillation\cite{gou2021knowledge} to enhance model performance without adding computational overhead. Initially, we train an iterative optimization-based stereo matching method\cite{wang2024selective} with sparse ground truth, using it as a teacher to generate dense pseudo-labels. These labels are then used to supervise the training of our model, followed by fine-tuning with the original sparse ground truth. This approach allows our model to learn richer and more detailed disparity information, improves the model's robustness, and ensures strong performance even with sparse annotations.

We conduct benchmark experiments on two distinct thermal stereo matching datasets\cite{shin2023MS2, treible2017cats}, covering daytime, nighttime, and rainy environments, including both indoor and outdoor scenes. The results show that our method achieves high precision in disparity estimation while ensuring real-time performance and excellent robustness. Our key contributions are summarized as follows:

\begin{itemize}
\item We propose ThermoStereoRT, a novel real-time thermal stereo matching network with lightweight components, achieving state-of-the-art accuracy and inference speed. 
\item We employ knowledge distillation to enhance disparity estimation without adding computational overhead, effectively overcoming the challenges of sparse ground truth and limited datasets.
\item We retrain existing stereo matching methods on thermal datasets, providing a comprehensive set of experiments that establish a new benchmark for accuracy and speed across various thermal stereo datasets.
\end{itemize}

\begin{figure*}[!h]
   \begin{center}
   \subfigure{
   \includegraphics[width=0.86\textwidth]{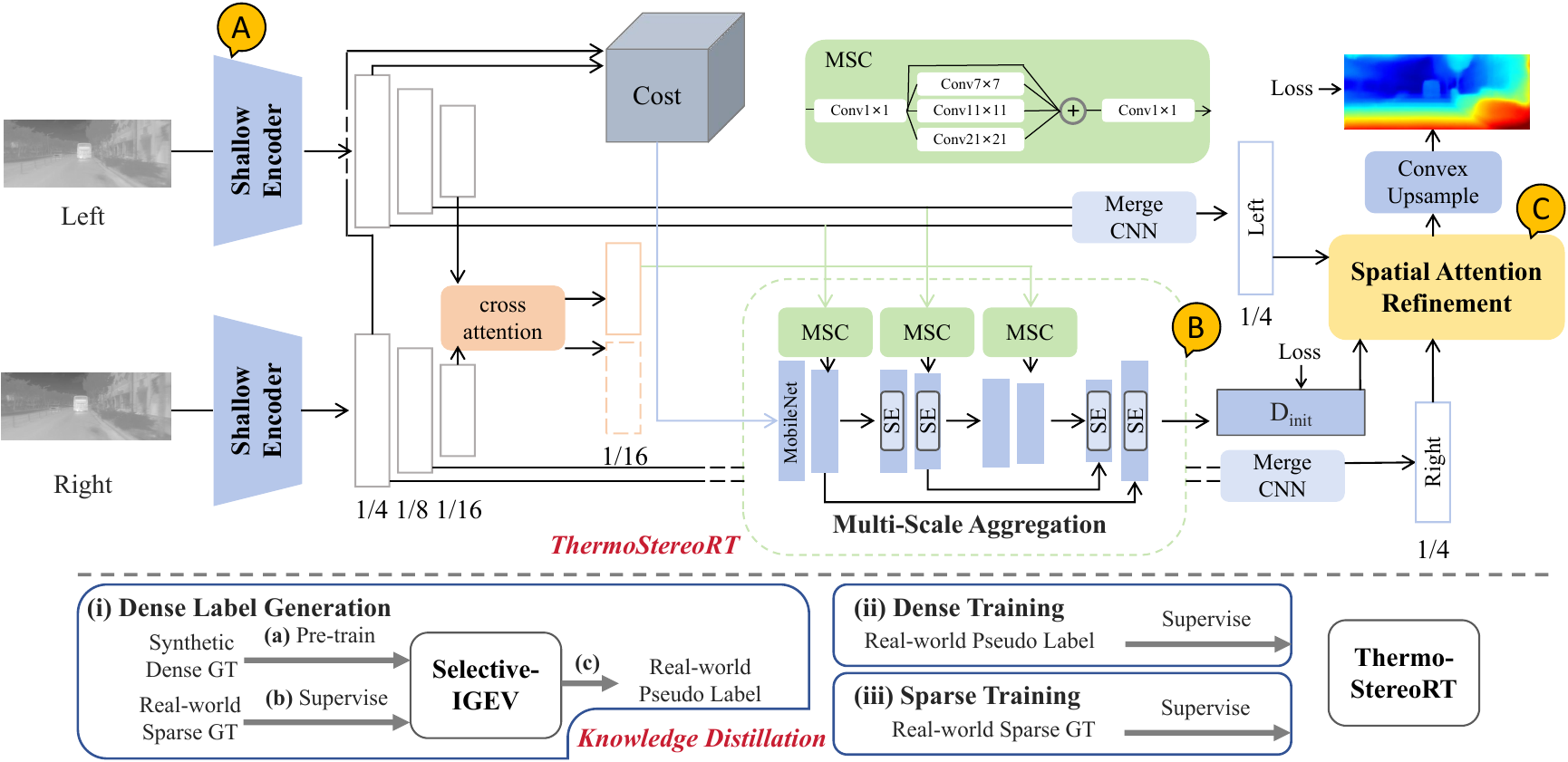}}

   \vspace{-0.3cm}
   \end{center}

\caption{Overview of our proposed ThermoStereoRT. First, stereo thermal images are fed into \textbf{A.} (shallow Encoder)  to generate features at different scales and construct a cost volume. Subsequently, \textbf{B.} (Multi-Scale Aggregation module) aggregates the cost and utilizes information from different scales. The initial disparity, along with the merged left and right features, is then fed into \textbf{C.} (Spatial Attention Refinement module) to refine details. The lower part of the figure illustrates the knowledge distillation process, where the Selective-IGEV\cite{wang2024selective} acts as the teacher for our work.}
   \label{fig:overview}
   \vspace{-0.3cm}
   \end{figure*}
   \vspace{-0.4cm}
\section{Related Work}
\subsection{Stereo Matching}

Stereo matching is a core challenge in robotic vision, aiming to estimate dense disparity maps from pairs of rectified RGB images. In recent years, the use of end-to-end neural networks has become the mainstream paradigm. To enhance the representational capacity of the cost volume, learning-based methods\cite{kendall2017end,guo2019group} typically employ CNN features to construct the cost volume, followed by 3D convolutions for its regularization.
To address the ambiguity issues in occluded regions and large texture-less regions, Chang et al.\cite{chang2018pyramid} and Guo et al.\cite{guo2019group} utilized 3D convolutions to regularize and filter the cost volume. However, the high computational complexity and memory consumption of 3D CNNs tend to hinder the application of these methods in high-resolution cost volumes. To improve efficiency, Shen et al.\cite{shen2021cfnet} introduce cascade method typically built a cost volume pyramid in a coarse-to-fine manner, progressively narrowing the disparity hypothesis range. Recently, iterative methods like RAFT-Stereo\cite{lipson2021raft} and CRE-Stereo\cite{li2022practical} have been proposed and achieved remarkable results, which recurrently update the disparity estimation using the local cost volume sampled from the all-pairs correlations. More recently, methods such as LightStereo\cite{guo2024lightstereo} and Selective-Stere\cite{wang2024selective} explored channel and spatial attention maps to regularize features or cost volumes, thus enhancing the network’s ability to perceive different regions of the image.

Nevertheless, in real-world applications, RGB image-based stereo matching methods\cite{lipson2021raft,xu2020aanet} can easily suffer from performance degradation under different weather conditions and dark environments. 
 In contrast, thermal images are not sensitive to different weather and light conditions. We are motivated to incorporate thermal images into the stereo matching architectures to improve its performance in these difficult environments.
\subsection{Thermal-based Stereo Matching}
Recent advancements in multi-modal stereo matching have included the integration of thermal imaging alongside traditional visible spectrum data. Liang et al.\cite{liang2022deepcrossspectral} proposes a deep cross-spectral stereo matching method to bridge the gap between RGB and NIR images through unsupervised learning, Liu et al.\cite{liu2022multi_thermal_visible} introduces a large-scale multi-view thermal-visible image dataset to facilitate cross-spectral matching in low-light conditions and proposes a semi-automatic approach for generating accurate supervision. Thermal-visible stereo matching improves accuracy in challenging conditions where RGB cameras might fail. However, matching between cross-spectral images remains challenging and difficult to implement practically.

CATS \cite{treible2017cats} is a Color and Thermal Stereo Benchmark; however, it lacks sufficient training data and includes some outdated models. MS2\cite{shin2023MS2} provides large-scale multimodal data including thermal stereo pairs for driving scenarios, but it is not specifically designed for thermal stereo matching algorithms, and the demonstrated results show significant room for improvement. There has been a lack of learning-based thermal stereo matching work in recent years. We aim for ThermoStereoRT to bridge this gap by providing a real-time network that addresses the inherent limitations of thermal images, such as lower resolution and lack of texture, paving the way for robust stereo matching systems applicable in diverse scenarios, including intelligent transportation and autonomous vehicles.


\section{Method}

Given a pair of rectified thermal images \( I_L\in \mathbb{R}^{H \times W} \) and \( I_R\in \mathbb{R}^{H \times W} \), our goal is to estimate the corresponding disparity map \( D_{\text{final}}\in \mathbb{R}^{H \times W} \) for the left image. Fig.\ref{fig:overview} illustrates the overall framework of ThermoStereoRT, which consists of three parts: (1) a shallow Encoder: This component efficiently extracts multi-scale features from stereo thermal images at resolutions of 1/4, 1/8, and 1/16. The features at the 1/4 resolution are used to construct the cost volume.
(2) an aggregation module: Based on residual connections and Squeeze-and-Excitation (SE) modules, this module utilizes channel boosting mechanisms and multi-scale attention to fully exploit the cost volume. (3) a refinement module based on spatial attention: This module leverages both local and global information from the stereo features to refine the disparity map. Due to the limited availability of stereo thermal imaging data and the sparsity of ground truth generated by LiDAR, the capabilities of stereo thermal matching models are constrained. We employ knowledge distillation techniques to enhance the performance of the stereo matching algorithm without introducing additional computational overhead.
\subsection{Shallow Encoder}
\begin{figure}[!h]
   \begin{center}
   \includegraphics[width=0.45\textwidth]{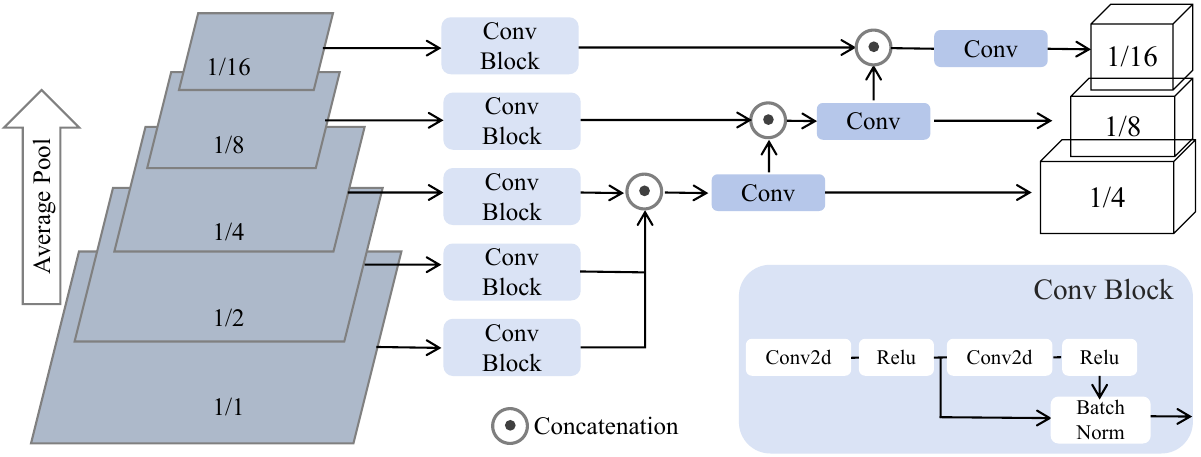}
   \end{center}
   \caption{Detailed architecture of the shallow encoder}
   \label{fig:TSRT_backbone}
   \vspace{-0.3cm}
   \end{figure}
Existing stereo matching algorithms often use deep CNNs or complex transformers for feature extraction, which limits the efficiency of the models. Inspired by NeuFlow\cite{zhang2024neuflow}, which uses a shallow CNN to extract features at 1/8 and 1/16 resolutions for optical flow tasks, we have proved in this work that for the task of thermal stereo matching, a shallow CNN is sufficient to extract multi-scale features.

As shown in the Fig. \ref{fig:TSRT_backbone}, given a thermal image \( I_M \), where \( M \in \left\{ \text{Left, Right}\right\} \), we first construct a thermal image pyramid using average pooling: \( P_{M,s} \) for \( s = 1, 2, 4, 8, 16 \), where \( P_{M,s} \in \mathbb{R}^{H/s \times W/s} \). To retain more original image information, we use \( P_{M,1}, P_{M,2}, P_{M,4} \) to extract 1/4 resolution features \( F_{M,4} \in \mathbb{R}^{{N_c} \times H/4 \times W/4 }\), where $N_c$ represents the number of channels. This process uses a convolution block as shown in the figure, containing only two convolution functions. High-resolution features are concatenated with lower-resolution features after downsampling, thus extracting better low-resolution features \( F_{M,8}, F_{M,16}  \). We construct a 3D correlation cost volume for each disparity level: 
\begin{equation}
\small
C_{c o r r}(d, x, y)=\frac{1}{N_c}\left\langle F_{L,4}(x, y), F_{R,4}(x-d, y)\right\rangle
\end{equation}
where $\langle\cdot, \cdot\rangle$ is the inner product of two vectors and \( C_{c o r r} \in \mathbb{R}^{ {D_{max}} \times H/4 \times W/4 }\), where $D_{max}$ is the max disparity.
This design makes the feature extraction module very lightweight and captures multi-scale features, which is beneficial for matching fine structures and handling large disparities.

\begin{figure*}[t]
\centering
\vspace{-0.5cm}
{\includegraphics[width=0.8\textwidth]{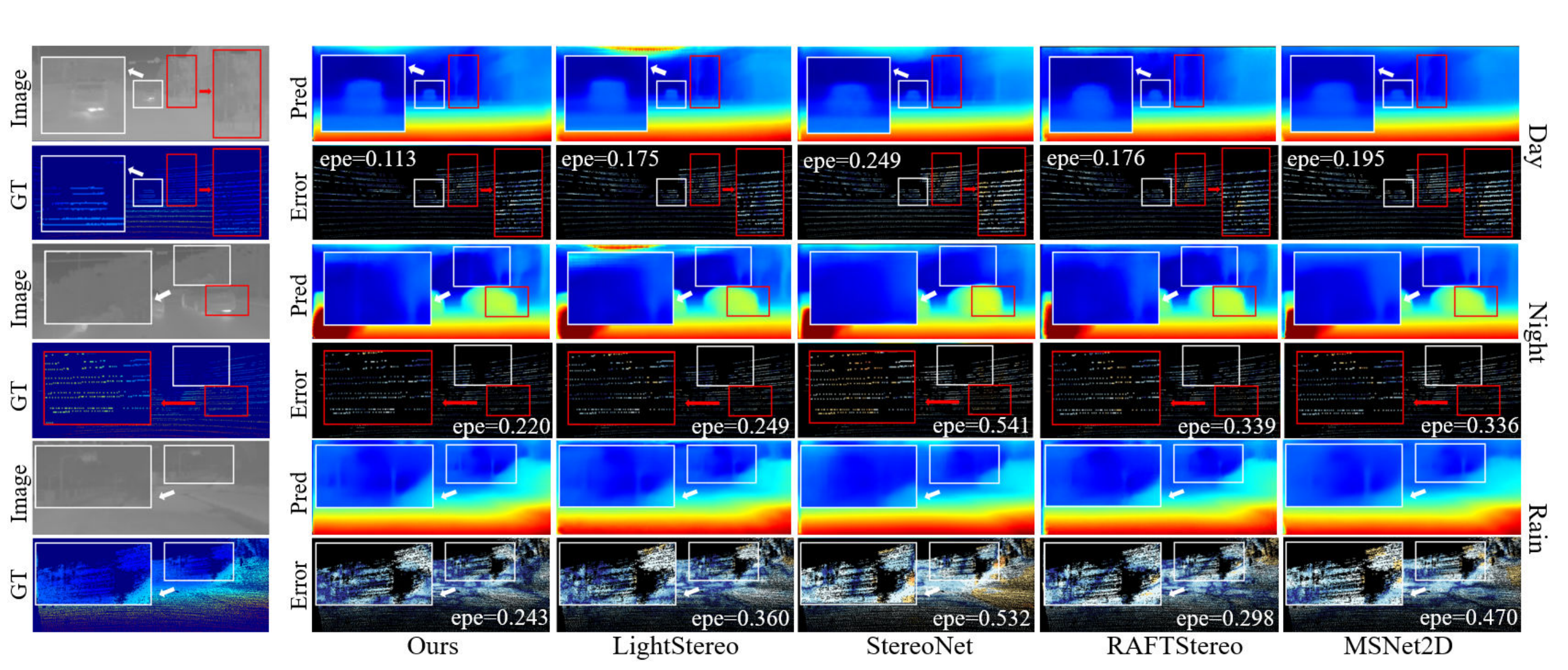}} 
\vspace{-0.3cm}
\caption{Qualitative results on MS2\cite{shin2023MS2} dataset. 
Our method is capable of predicting fine disparity from thermal images with a small error (blue in error map).}
\label{fig:ms2}
\vspace{-0.3cm}
\end{figure*}
\subsection{Multi-Scale Aggregation}
As shown in Fig.\ref{fig:overview}, we utilize MobileNetV3\cite{koonce2021mobilenetv3} residual blocks to construct a 3D aggregation network. To retain details and facilitate gradient propagation, we incorporate residual connections at 1/4 and 1/8 resolutions. 
To compensate for the information loss inherent in the process of building the correlation cost volume, we employ multi-scale convolutions (MSC) at three feature maps \( F_{L,4}, F_{L,8}, \) and \( F_{\text{cross},16} \). Here, \( F_{\text{cross},16} \) is obtained by using \( F_{L,16} \) as the query and \( F_{R,16} \) as the key and value, through global cross-attention. This operation enhances the feature distinctiveness of \( F_{\text{cross},16} \). The MSC comprises convolutions of sizes 1x1, 7x7, 11x11, and 21x21, which capture both local and global information within the feature maps. The output of the MSC serves as attention weights, which are multiplied with the intermediate outputs of the aggregation network. These blocks aggregate features from neighboring disparities and pixels to predict refined cost volumes $C_{\text{refine}}$.
\begin{equation}\small
C_{\text {refine }}=\operatorname{Aggregation}(C_{\text {corr }}, \operatorname{MSC}\left(F_{L,4}, F_{L,8}, F_{\text {cross}, 16}\right))
\end{equation}
We utilize derivable disparity regression to estimate the continuous disparity map. The predicted disparity \( D_{\text {init }} \) is computed by the soft argmin function:

\begin{equation}
\small
D_{\text {init }} = \sum_{d=0}^{D_{\text{max}}} d \times \sigma(C_{\text {refine }})
\end{equation}
where the probability of each disparity \( d \) is calculated from the predicted cost \( C_{\text {refine }} \) via the softmax operation \( \sigma(\cdot) \).

\begin{table*}[htbp]
   \caption{Results of benchmark tests on MS2\cite{shin2023MS2}.
   }
\vspace{-0.5cm}

   \begin{center}
      \scriptsize
   \resizebox{0.95\textwidth}{!}{ \begin{minipage}{\textwidth} 
   \begin{tabular}{clccccccccccccc}
   \toprule
   \multicolumn{1}{c}{\multirow{2}{*}{10FPS}}  &\multicolumn{1}{l}{\multirow{2}{*}{Model}}  & \multicolumn{3}{c}{Day} & \multicolumn{3}{c}{Night} &  \multicolumn{3}{c}{Rain} &\multicolumn{1}{c}{FLOPs} &\multicolumn{1}{c}{ Params} &\multicolumn{2}{c}{ FPS(Hz)}\\                                         
                                                                                    \cmidrule(r){3-5}                  \cmidrule(r){6-8} \cmidrule(r){9-11}    \cmidrule(r){14-15}
   \multicolumn{1}{c}{Jetson}   &     \multicolumn{1}{c}{}            & \multicolumn{1}{c}{EPE $\downarrow$}                     & \multicolumn{1}{c}{$>0.5$ $\downarrow$} & \multicolumn{1}{c}{$>1$ $\downarrow$} & \multicolumn{1}{c}{EPE $\downarrow$}                     & \multicolumn{1}{c}{$>0.5$ $\downarrow$} & \multicolumn{1}{c}{$>1$ $\downarrow$} & \multicolumn{1}{c}{EPE $\downarrow$}                     & \multicolumn{1}{c}{$>0.5$ $\downarrow$} & \multicolumn{1}{c}{$>1$ $\downarrow$}& \multicolumn{1}{c}{(G)} & \multicolumn{1}{c}{(M)}  & \multicolumn{1}{c}{A6000}& \multicolumn{1}{c}{Jetson}   \\
   \midrule
   \multicolumn{1}{c}{\multirow{8}{*}{\rotatebox{90}{Slower}}} & PSMNet\cite{chang2018pyramid}     & \underline{0.2133} & 7.970  & 1.977  & 0.3441 & 19.687 & 6.685  & 0.2798  &\underline{12.899}   & \underline{3.179} & 155.53  & 5.22  &34.32  &2.34 \\
       \multicolumn{1}{c}{}& AANet\cite{xu2020aanet}    &0.3684  &18.569  & 6.077   &0.5977  &36.524 &16.534 &0.4608  &25.709  &9.107  &\underline{32.41} &2.70  &43.86&6.57   \\
   \multicolumn{1}{c}{}& Mocha\cite{chen2024mocha}    & 0.2670 & 11.771&  3.111 & 0.3863  & 23.169 & 8.0130 & 0.3317  & 17.163  &  4.608& 615.52 &20.75  & 8.34 &0.65  \\   
   \multicolumn{1}{c}{}& RAFT-Stereo\cite{lipson2021raft}    &0.2937  &12.739   & 3.571  &0.4282   &25.998 &9.767  &0.3544 &19.036   &5.427     &467.39  &11.10&14.90  &0.93   \\

   \multicolumn{1}{c}{}& Selective-IGEV\cite{wang2024selective}    & \textbf{0.1950} & \textbf{6.428} & \textbf{1.553}  & \textbf{0.2904}  & \textbf{15.598} & \textbf{4.649} &  \textbf{0.2588} &  \textbf{11.304} &  \textbf{2.631} & 501.30 & 13.14 & 13.43  &0.89  \\


   \multicolumn{1}{c}{}& MSNet2D\cite{shamsafar2022mobilestereonet}   &0.3332  &15.085   & 3.928 &0.4416   &26.032 &9.138 & 0.3920  &22.060   &6.254   &41.36  &\underline{2.35}  &28.33   &3.76  \\
    \multicolumn{1}{c}{}& MSNet3D\cite{shamsafar2022mobilestereonet}   &0.2137  &\underline{7.754}   & \underline{1.878}  &0.3167   &17.684  &5.508&\underline{0.2777}  &13.181  &3.292  &70.35  &\textbf{1.86} &\underline{54.25} &\underline{8.40}  \\
    \multicolumn{1}{c}{}& Fast-ACVNet\cite{xu2023accurate}    &0.2143 &8.263   & 1.986  &\underline{0.3114}   &\underline{17.351}  &\underline{5.241} &0.2907  &13.960 &3.463  &\textbf{19.43}  &3.08  &\textbf{69.49} &\textbf{9.33}   \\

       \midrule

    \multicolumn{1}{c}{\multirow{6}{*}{\rotatebox{90}{Faster}}}& StereoNet\cite{khamis2018stereonet}    &0.4615  &27.445 &9.887  &0.6448  &39.183 &18.226 &0.6010  &36.374  &14.288 &\textbf{4.23 }&\textbf{0.76 } &\textbf{189.12 }&\textbf{22.20}  \\
        \multicolumn{1}{c}{}& LightStereo\cite{guo2024lightstereo}    &0.2535  &10.344 &2.588  &\underline{0.3607}  &21.303 &\underline{6.829} &0.3210  &16.564&4.382 &\underline{4.65}&\underline{2.07}  &102.71 &14.03  \\

      \multicolumn{1}{c}{}& Ours-T    & 0.2676&  11.639&  3.009 & 0.4116 & 25.151& 9.029 & 0.3356 & 17.636 &  4.770 & 24.29 &2.40 & \underline{147.40} &\underline{19.90} \\   
         \multicolumn{1}{c}{}& Ours-S   & \underline{0.2297} &  \underline{9.137} &  \underline{2.248} & 0.3681 & 21.842 & 7.358 &\underline{0.2984}  & \underline{14.621} &  \underline{3.741} &31.37 &3.09  & 118.45  &14.79\\   
         \multicolumn{1}{c}{} & Ours w/o KD    & 0.2405 & 9.688  & 2.564  & 0.3680  & \underline{21.141} & 7.324  & 0.3152  & 15.187 & 4.110 & 31.38  & 3.21  & 106.23  &12.58 \\
   \multicolumn{1}{c}{}& Ours    & \textbf{0.2240} &  \textbf{8.727} &  \textbf{2.142} & \textbf{0.3426}  & \textbf{19.567} & \textbf{6.404} & \textbf{0.2934}  & \textbf{13.861}  &  \textbf{3.500} & 31.38 &3.21  & 105.89 &12.58 \\   
   \midrule
   
   \end{tabular}
   
   \end{minipage}}

   \label{tab:results for MS2}
      \end{center}    
\vspace{-0.5cm}
   \end{table*}

\subsection{Spatial Attention Refinement}
\begin{figure}[!h]
   \begin{center}
   \includegraphics[width=0.45\textwidth]{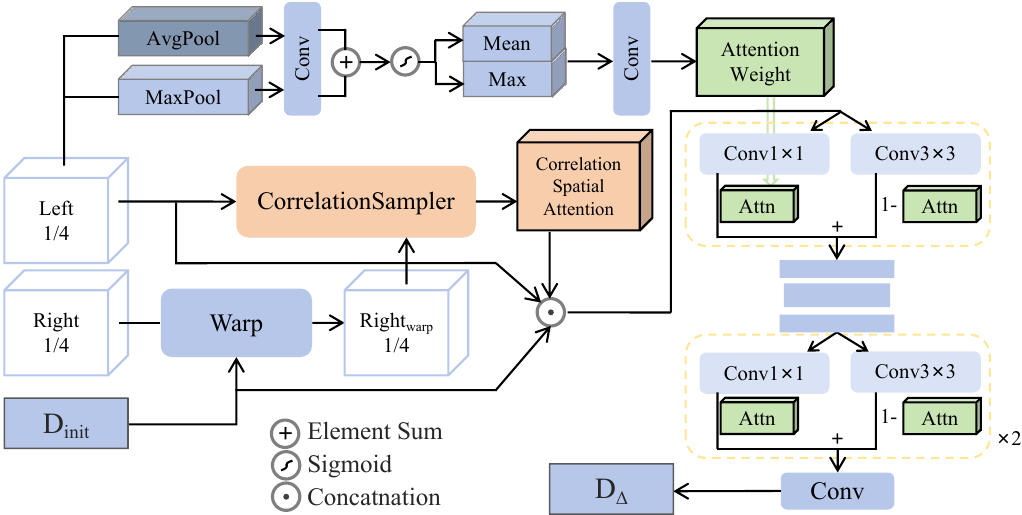}
   \end{center}
   \vspace{-0.4cm}
   \caption{Spatial attention refinement module. The module generate detailed disparity adjustment from initial disparity and merged features.}
   \label{fig:TSRT_refine}
    \vspace{-0.4cm}
   \end{figure}

We have designed a lightweight spatial attention refinement module to estimate fine disparity adjustments, as shown in Fig. \ref{fig:TSRT_refine}. Many stereo matching algorithms based on iterative optimization repeatedly estimate disparity adjustments to refine the initial disparity, which can be time-consuming. Our approach involves a spatial attention-based refinement algorithm that uses attention weights to modulate both small and large kernels, thereby expanding the receptive field. This allows us to refine the initial disparity in a single operation rather than iteration.

The input to the refinement module is the merged features from the 1/4 and 1/8 resolutions:
\begin{equation} \small
\begin{split}
F_{L,Merge} = \text{MergeCNN}(F_{L,4}, F_{L,8}), \\
F_{R,Merge} = \text{MergeCNN}(F_{R,4}, F_{R,8})
\end{split}
\end{equation}
The merged right feature is warped to the left viewpoint based on the initial disparity. The warped features are then correlated with the left-image features to compute the correlation spatial attention $W_{\text{corr}}$, which reflects the misalignment of the warped features and aids in optimizing the disparity adjustment. Subsequently, the correlation spatial attention is concatenated with $F_{L,Merge} \text{ and } D_{\text{init}}$ to generate $F_{\text{Concat}}$, as follows:
\begin{equation}
\small
W_{\text{corr}} = \operatorname{Correlation}\left(F_{L,Merge}, \operatorname{Warp}\left(F_{R,Merge}, D_{\text{init}}\right)\right)
\end{equation}
\begin{equation}
\small
F_{\text{Concat}} = \operatorname{Concat}\left(F_{L,Merge}, D_{\text{init}}, W_{\text{corr}} \right)
\end{equation}
To integrate information from different receptive fields and retain more details, we use 1x1 and 3x3 convolutional kernels for decoding the disparity adjustment, as shown in the right side of Fig. \ref{fig:TSRT_refine}. Unlike directly concatenating the outputs of these two convolutions, we train an attention weight $W_{\text{attn}}$ based on global information and channel attention to adaptively merge the outputs. This design adds minimal computational overhead but significantly improves model performance compared to using only 3x3 convolutional kernels for decoding. It better recovers fine edge details and semantic information. This process can be formulated as follows:
\begin{equation}\small
    D_{\Delta} = \operatorname{ConvLayers}(W_{\text{attn}}, F_{\text{Concat}})
\end{equation}
After the spatial attention refinement, the disparity adjustment is added to the initial disparity. The result is then upsampled to the image resolution using the convex upsampling module from RAFT\cite{teed2020raft}, preserving finer details and getting the final disparity.
\begin{equation}\small
    D_{\text{final}} = \operatorname{Upsample}(D_{\text{init}} +D_{\Delta} )
\end{equation}

\subsection{Knowledge Distillation for ThermoStereoRT}
Thermal stereo data typically has a lower resolution, exacerbating the impact of sparse ground truth and necessitating advanced techniques like knowledge distillation to improve model performance. Inspired by DepthAnythingv2\cite{yang2024depthanythingv2}, which highlights the impact of dense versus sparse ground truths, we leverage knowledge distillation to bridge the gap between sparse and dense data, enhancing the robustness and detail fidelity of our model despite the limitations of thermal datasets.

The knowledge distillation process integrated into our ThermoStereoRT consists of three main stages: (1)Dense Label Generation, (2)Dense Training, and (3)Sparse Training. (1) we pre-train a computationally intensive Selective-IGEV network using synthetic data and then train it with sparse ground truth to generate high-quality dense labels. (2)These dense pseudo labels, produced by the Selective-IGEV model, are then used to supervise the training of ThermoStereoRT via knowledge distillation, enhancing performance without increasing computational demands. (3)The ThermoStereoRT model is fine-tuned with original sparse ground truth to optimize performance. We choose Selective-IGEV as the teacher model for knowledge distillation, because this method achieved low EPE values in the same experimental setting as all models.

\subsection{Loss Function}
We supervise benchmark training and knowledge distillation using a sequence loss defined as the L1 distance between predicted and ground truth disparities, with exponentially increasing weights.

Given the ground truth disparity \( D_{\text{gt}} \), the loss \( L \) is:
\begin{equation} \small
    L = \sum_{i=1}^{N} \gamma^{N-i} \| D_{\text{gt}} - D_i \|_1
\end{equation}
where \( \gamma = 0.9 \) and \( N \) is the number of predictions in the sequence. 
When training ThermoStereoRT, the outputs are \( D_{\text{init}} \) and \( D_{\text{final}} \), so \( N = 2 \). This loss is applied to all models during benchmark training of thermal stereo matching.

\section{Experiments}
\subsection{Datasets}

We conducted benchmark testing on the MS2\cite{shin2023MS2} and CATS\cite{treible2017cats} datasets. The MS2\cite{shin2023MS2} (Multi-Spectral Stereo) dataset comprises approximately 195,000 synchronized and rectified multi-modal data pairs, collected from different scenarios, covering different times of day and weather conditions. We utilized 76,544 thermal stereo image pairs for training, 400 pairs for validation, and an additional 23,316, 22,915, and 25,022 pairs for testing under daytime, nighttime, and rainy conditions respectively. The resolution of the thermal images is 256 $\times$ 640.

The CATS\cite{treible2017cats} (Color And Thermal Stereo) dataset includes around 1,400 images covering cluttered indoor and outdoor scenes, featuring challenging environments and conditions. However, CATS\cite{treible2017cats} contains a relatively small number of thermal image pairs. We split the dataset using 80 pairs of thermal images for indoor scenes for training and 20 pairs for validation, while for outdoor scenes, 54 pairs are used for training and 14 pairs for validation. The resolution of these thermal images is 480 $\times$ 640.

The SceneFlow dataset consists of over 39,000 synthetic stereo RGB image pairs, with 34,801 training image pairs having precise ground truth disparity. The image size of SceneFlow is 540 $\times$ 960. We converted the SceneFlow dataset into grayscale and scaled the pixel values to a range of 0-40 to make the grayscale values more closely resemble those of thermal images.
\subsection{Implementation Details}

ThermoStereoRT is implemented using PyTorch\cite{paszke2019pytorch} and is trained on a single NVIDIA A6000 GPU. When training on thermal image datasets, we maintain the original resolution with a batch size of 4, utilizing the AdamW optimizer and employing a one-cycle learning rate schedule with a maximum learning rate of 0.001. For the MS2\cite{shin2023MS2} dataset, all methods are trained for 200k steps. For the CATS\cite{treible2017cats} dataset, all methods undergo 30k steps of training.
For knowledge distillation, the teacher model is first trained for 100k steps on the grayscale version of the SceneFlow dataset, followed by another 150k steps on the MS2\cite{shin2023MS2} dataset. The student model is initially trained for 100k steps under the supervision of the teacher model and then further trained for 150k steps on the MS2\cite{shin2023MS2} dataset. For the CATS\cite{treible2017cats} dataset, the distillation process involves the teacher supervising the student for 30k steps, after which the student continues training for another 30k steps independently.
Given the inherently lower resolution of thermal image datasets and the fact that temperature values carry practical significance, we refrained from applying extensive data augmentation post knowledge distillation to preserve the integrity and meaningfulness of the thermal data.

\begin{table}[ht]
   \caption{Results of benchmark tests on CATS\cite{treible2017cats}. }
   \begin{center}
   \vspace{-0.4cm}
      \scriptsize
   \resizebox{0.95\textwidth}{!}{ \begin{minipage}{\textwidth} 
   \begin{tabular}{lccccccc}
   \toprule
 \multicolumn{1}{l}{\multirow{2}{*}{Model}}&     \multicolumn{1}{c}{}  & \multicolumn{3}{c}{Indoor} & \multicolumn{3}{c}{Outdoor} \\                                         
                                                                                    \cmidrule(r){3-5}                  \cmidrule(r){6-8}
   \multicolumn{1}{c}{}   &     \multicolumn{1}{c}{}            & \multicolumn{1}{c}{EPE $\downarrow$}                     & \multicolumn{1}{c}{$>1$ $\downarrow$} & \multicolumn{1}{c}{$>5$ $\downarrow$} & \multicolumn{1}{c}{EPE $\downarrow$}                     & \multicolumn{1}{c}{$>1$ $\downarrow$} & \multicolumn{1}{c}{$>5$ $\downarrow$}  \\
   \midrule
   PSMNet\cite{chang2018pyramid}     & \multicolumn{1}{c}{} &1.119 &29.30& 4.268  & 0.9773 &27.13 & 3.103  \\
   RAFTStereo\cite{lipson2021raft}     & \multicolumn{1}{c}{} & 0.9885  & 26.11 &3.690 & 1.002 &29.06 & 2.916  \\
   StereoNet\cite{khamis2018stereonet}    & \multicolumn{1}{c}{} & 1.460  & 36.13 & 6.988  & 1.694 &46.95 & 9.125 \\
   LightStereo\cite{guo2024lightstereo}     & \multicolumn{1}{c}{} & 1.240  &29.39 & 6.297 & 1.209 &27.99 &6.281  \\
    Ours     & \multicolumn{1}{c}{} & 1.074  & 28.17 & 3.960  & 1.052 &28.62 & 3.419  \\

   \midrule
   
   \end{tabular}
   
   \end{minipage}}

   \label{tab:results for CATS}
      \end{center}    
\vspace{-0.8cm}
   \end{table}

\begin{table}[ht]
   \caption{Performance gains from knowledge distillation
   }
   \vspace{-0.4cm}
   \begin{center}
      \scriptsize
   \resizebox{0.48\textwidth}{!}{
   \begin{tabular}{lccccccc}
   \toprule
 \multicolumn{1}{l}{\multirow{2}{*}{Model}}&  \multicolumn{1}{c}{}  & \multicolumn{1}{c}{Day} & \multicolumn{1}{c}{Night}& \multicolumn{1}{c}{Rain} &\multicolumn{3}{c}{Performance Gain} \\                                         
  \cmidrule(r){3-3} \cmidrule(r){4-4}\cmidrule(r){5-5}  \cmidrule(r){6-8}
   \multicolumn{1}{c}{}   &     \multicolumn{1}{c}{}            & \multicolumn{1}{c}{EPE $\downarrow$}                     & \multicolumn{1}{c}{EPE $\downarrow$} & \multicolumn{1}{c}{EPE $\downarrow$} & \multicolumn{1}{c}{Day}                     & \multicolumn{1}{c}{Night} & \multicolumn{1}{c}{Rain}  \\
      \midrule
   LightStereo     & \multicolumn{1}{c}{} & 0.2535  & 0.3608  & 0.3210  & \multicolumn{1}{c}{\multirow{2}{*}{7.25\%}} & \multicolumn{1}{c}{\multirow{2}{*}{5.63\%}} & \multicolumn{1}{c}{\multirow{2}{*}{4.14\%}} \\
LightStereo+KD     & \multicolumn{1}{c}{} & 0.2351  & 0.3405 & 0.3077  &  &   &   \\
   \midrule
    Ours w/o KD     & \multicolumn{1}{c}{} & 0.2405  & 0.3677  & 0.3152 &  \multicolumn{1}{c}{\multirow{2}{*}{6.82\%}} & \multicolumn{1}{c}{\multirow{2}{*}{6.83\%}} & \multicolumn{1}{c}{\multirow{2}{*}{6.92\%}}  \\
Ours     & \multicolumn{1}{c}{} & 0.2241 & 0.3426 & 0.2934  &  &   &  \\

   \midrule
       Ours-S w/o KD     & \multicolumn{1}{c}{} & 0.2563  & 0.3890  & 0.3124 & \multicolumn{1}{c}{\multirow{2}{*}{10.38\%}} & \multicolumn{1}{c}{\multirow{2}{*}{5.37\%}} & \multicolumn{1}{c}{\multirow{2}{*}{4.48\%}} \\
Ours-S    & \multicolumn{1}{c}{} & 0.2297  &0.3681 &0.2984  &  &   &   \\

   \midrule
   
       Ours-T w/o KD    & \multicolumn{1}{c}{} & 0.2944  & 0.4439  & 0.3633  & \multicolumn{1}{c}{\multirow{2}{*}{9.07\%}} & \multicolumn{1}{c}{\multirow{2}{*}{7.28\%}} & \multicolumn{1}{c}{\multirow{2}{*}{7.62\%}} \\
Ours-T     & \multicolumn{1}{c}{} & 0.2677 &0.4116  &0.3356  &  &   &   \\
   \midrule
   \end{tabular}
}

   \label{tab:KD}
      \end{center}    
\vspace{-0.8cm}
   \end{table}

\begin{table}[ht]
   \caption{Ablation study
   }
   \vspace{-0.8cm}
   \begin{center}
      \scriptsize
   \resizebox{0.95\textwidth}{!}{ \begin{minipage}{\textwidth} 
   \begin{tabular}{lccccccc}
   \toprule
 \multicolumn{1}{l}{\multirow{2}{*}{Model}}&  \multicolumn{1}{c}{}  & \multicolumn{2}{c}{Day} & \multicolumn{2}{c}{Night}& \multicolumn{2}{c}{Rain}\\                                         
  \cmidrule(r){3-4} \cmidrule(r){5-6}\cmidrule(r){7-8}
   \multicolumn{1}{c}{}   &     \multicolumn{1}{c}{}           & \multicolumn{1}{c}{EPE $\downarrow$}                     & \multicolumn{1}{c}{$>0.5$ $\downarrow$}                   & \multicolumn{1}{c}{EPE $\downarrow$}                     & \multicolumn{1}{c}{$>0.5$ $\downarrow$} & \multicolumn{1}{c}{EPE $\downarrow$}                     & \multicolumn{1}{c}{$>0.5$ $\downarrow$}  \\
   \midrule
    Ours     & \multicolumn{1}{c}{} & 0.2405 & 9.688  & 0.3680  & 21.141& 0.3152  & 15.187 \\
    w/o $W_{attn}$    & \multicolumn{1}{c}{} & 0.2607  & 10.791  & 0.3839  & 21.944 &0.3321   &16.601 \\
    w/o $W_{corr}$    & \multicolumn{1}{c}{} & 0.2556  & 10.844  & 0.3824  & 22.533 &0.3269  &16.504 \\
w/o SE    & \multicolumn{1}{c}{} & 0.2563  & 10.306 & 0.3890 & 23.236 &0.3124   &15.669 \\
w/o refine  & \multicolumn{1}{c}{} & 0.2847  & 12.558 & 0.4099  & 23.825&0.3625  &18.212 \\
w/o SE/refine & \multicolumn{1}{c}{} &0.2944 &13.370  & 0.4439 &27.081 &0.3633 &19.498  \\

   \midrule

   \end{tabular}
   
   \end{minipage}}

   \label{tab:results for ablation}
      \end{center}    
\vspace{-0.8cm}
   \end{table}






   



\subsection{Benchmark Evaluation}
We retrained and evaluated all methods designed for precision or efficiency on the MS2\cite{shin2023MS2} and CATS\cite{treible2017cats} datasets, providing a reliable benchmark for thermal stereo matching, Tab. \ref{tab:results for MS2} for MS2\cite{shin2023MS2}, and Tab. \ref{tab:results for CATS} for CATS\cite{treible2017cats}. 
We use the end point error(EPE) and percentage of disparity outliers (error $>$ n) to evaluate the methods. Tab. \ref{tab:results for MS2} shows our method achieves superior results across different environments while ensuring real-time performance. Iterative optimization methods like Mocha\cite{chen2024mocha} and RAFT-Stereo\cite{lipson2021raft} perform suboptimally when processing low-resolution thermal images, while Selective-IGEV\cite{wang2024selective} achieved low EPE values in the same experimental setting as all models. Methods relying on stacked 3D convolutions, such as PSMNet\cite{chang2018pyramid} and MSNet3D\cite{shamsafar2022mobilestereonet}, can achieve low EPE values in thermal scenarios; however, these approaches struggle to maintain accuracy when 3D convolutions are removed, making it difficult to ensure real-time performance. 
Our method achieves competitive EPE values compared to more resource-intensive methods. When contrasted with recent work such as LightStereo\cite{guo2024lightstereo}, ours not only operates faster on NVIDIA A6000 but also improves EPE performance by 11.6\%. Compared to StereoNet\cite{khamis2018stereonet}, ours shows a remarkable 51.5\% improvement in EPE value. Our method offers two additional variants: Ours-S, which omits the SE module in the aggregation module, and Ours-T, which excludes both the SE and refine modules, catering to scenarios with extremely high real-time requirements. Notably, Ours-S outperforms LightStereo\cite{guo2024lightstereo} both in terms of accuracy and speed.

Fig. \ref{fig:ms2} vividly demonstrates the superior performance of our method in outdoor driving scenarios. Our approach can recover excellent details from thermal stereo images, identifying distant vehicles, trees, and poles. Tab. \ref{tab:results for CATS} showcases the results of various methods on the CATS\cite{treible2017cats} dataset, highlighting that our method achieves the best EPE among real-time algorithms and exhibits a lower percentage of large pixel outliers. Fig. \ref{fig:cats} illustrates the performance of different real-time algorithms in typical indoor and outdoor settings. Our method successfully recovers balls from low-resolution, blurry indoor thermal images and achieves markedly clearer segmentation between objects. In outdoor scenarios, our method is able to recover human shape well, whereas other algorithms fail, demonstrating the robustness of our approach.

\subsection{Knowledge Distillation Performance}

Tab. \ref{tab:KD} shows the performance gains achieved by different models through knowledge distillation. These models are trained for 200k steps on the MS2\cite{shin2023MS2} dataset, ensuring convergence. During knowledge distillation, the models are first trained for 100k steps using pseudo labels, followed by an additional 150k steps on the MS2\cite{shin2023MS2} dataset. It is evident that knowledge distillation enabled the models to optimize to jump out of saddle points, delivering improved performance without increasing the computational load.
\subsection{Ablation Study}
To validate the effectiveness of different components in our method, we conducted a series of ablation studies on the MS2\cite{shin2023MS2} dataset without knowledge distillation in Tab. \ref{tab:results for ablation}. The attention-based refinement module plays a crucial role in recovering details and enhancing the model's generalizability; when this module is omitted, the performance of the model drops significantly. The specific designs within the refinement module are also essential. $W_{attn}$, generated by spatial attention and used to modulate the 1x1 convolutions, aids in extracting detailed information. Experiments show that removing $W_{attn}$ leads to a notable decrease in model accuracy. Similarly, $W_{corr}$ is vital for leveraging the information from both left and right features. While neither $W_{attn}$ nor $W_{corr}$ substantially increases the computational load, both contribute significantly to performance improvements. The SE\cite{hu2018squeeze} (Squeeze-and-Excitation) module, incorporated into the aggregation module based on MobileNetV3\cite{koonce2021mobilenetv3} residual convolutions, also contributes to improved model performance.

\subsection{Real-time Performance}

We evaluate the real-time performance of different algorithms on the NVIDIA Jetson Xavier NX which delivers up to 21 TOPS. As shown in Tab. \ref{tab:results for MS2}, Our method demonstrates outstanding real-time performance, with ours-S achieving nearly 15 frames per second, which is sufficient for many downstream tasks while exhibiting excellent performance. Algorithms like PSMNet\cite{chang2018pyramid} and Selective-IGEV\cite{wang2024selective} show good performance in our benchmark experiments but can not be executed on embedded devices. Our real-time performance enables seamless integration into resource-constrained environments, ensuring our solution deployable in real-world scenarios where computational efficiency is critical.
\section{Conclusion}
In this paper, we propose ThermoStereoRT, an advanced real-time thermal stereo matching method suitable for all-weather indoor and outdoor scenes. We design a lightweight encoder, utilizing multi-scale attention aggregation, and introduce a novel attention-based refinement module combining channel and spatial information. To tackle the limited availability and sparsity of ground truth data in thermal imagery, we use knowledge distillation to enhance performance without additional computation. Extensive testing demonstrates real-time processing and robust performance across multiple datasets and real-world scenarios. Deployable on mobile devices, ThermoStereoRT aims to contribute to the development of thermal stereo matching and establish a new benchmark.

{\small
\bibliographystyle{ieee_fullname}
\bibliography{ICRA_new_tmp}
}

\end{document}